\documentclass[10pt,twocolumn,letterpaper]{article}

\usepackage{cvpr}              
\definecolor{cvprblue}{rgb}{0.21,0.49,0.74}
\usepackage[pagebackref,breaklinks,colorlinks,allcolors=cvprblue]{hyperref}
\usepackage{multirow}
\usepackage{overpic}
\usepackage{pifont}
\usepackage{graphicx} 
\usepackage{tabularx}
\newcommand{\cmark}{\ding{51}}
\newcommand{\xmark}{\ding{55}}

\def\confName{CVPR}
\def\confYear{2026}

\title{TriLite: Efficient Weakly Supervised Object Localization with Universal Visual Features and Tri-Region Disentanglement}

\author{Arian Sabaghi, José Oramas \\
University of Antwerp, sqIRL/IDLab, imec\\
Antwerp, Belgium\\
}

\begin{document}
\maketitle
\begin{abstract}
Weakly supervised object localization (WSOL) aims to localize target objects in images using only image-level labels. 
Despite recent progress, many approaches still rely on multi-stage pipelines or full fine-tuning of large backbones, which increases training cost, while the broader WSOL community continues to face the challenge of partial object coverage. 
We present \textit{TriLite}, a single-stage WSOL framework that leverages a frozen Vision Transformer with Dinov2 pre-training in a self-supervised manner, and introduces only a minimal number of trainable parameters (fewer than 800K on ImageNet-1K) for both classification and localization. 
At its core is the proposed \textit{TriHead} module, which decomposes patch features into foreground, background, and ambiguous regions, thereby improving object coverage while suppressing spurious activations. 
By disentangling classification and localization objectives, TriLite effectively exploits the universal representations learned by self-supervised ViTs without requiring expensive end-to-end training. 
Extensive experiments on CUB-200-2011, ImageNet-1K, and OpenImages demonstrate that TriLite sets a new state of the art, while remaining significantly more parameter-efficient and easier to train than prior methods.
The code will be released soon.
\end{abstract}    
\section{Introduction}
\label{sec:intro}

\begin{figure*}
    \centering
    
    \begin{minipage}{0.485\linewidth}
        \centering
        \makebox[0.24\linewidth][c]{\small Image}%
        \makebox[0.24\linewidth][c]{\small CAM}%
        \makebox[0.24\linewidth][c]{\small GenPromp}%
        \makebox[0.24\linewidth][c]{\small Ours}
    \end{minipage}
    \hspace{-0.01\linewidth}
    \begin{minipage}{0.485\linewidth}
        \centering
        \makebox[0.24\linewidth][c]{\small Image}%
        \makebox[0.24\linewidth][c]{\small CAM}%
        \makebox[0.24\linewidth][c]{\small GenPromp}%
        \makebox[0.24\linewidth][c]{\small Ours}
    \end{minipage}

    \begin{minipage}{0.485\linewidth}
        \centering
        \includegraphics[width=\linewidth]{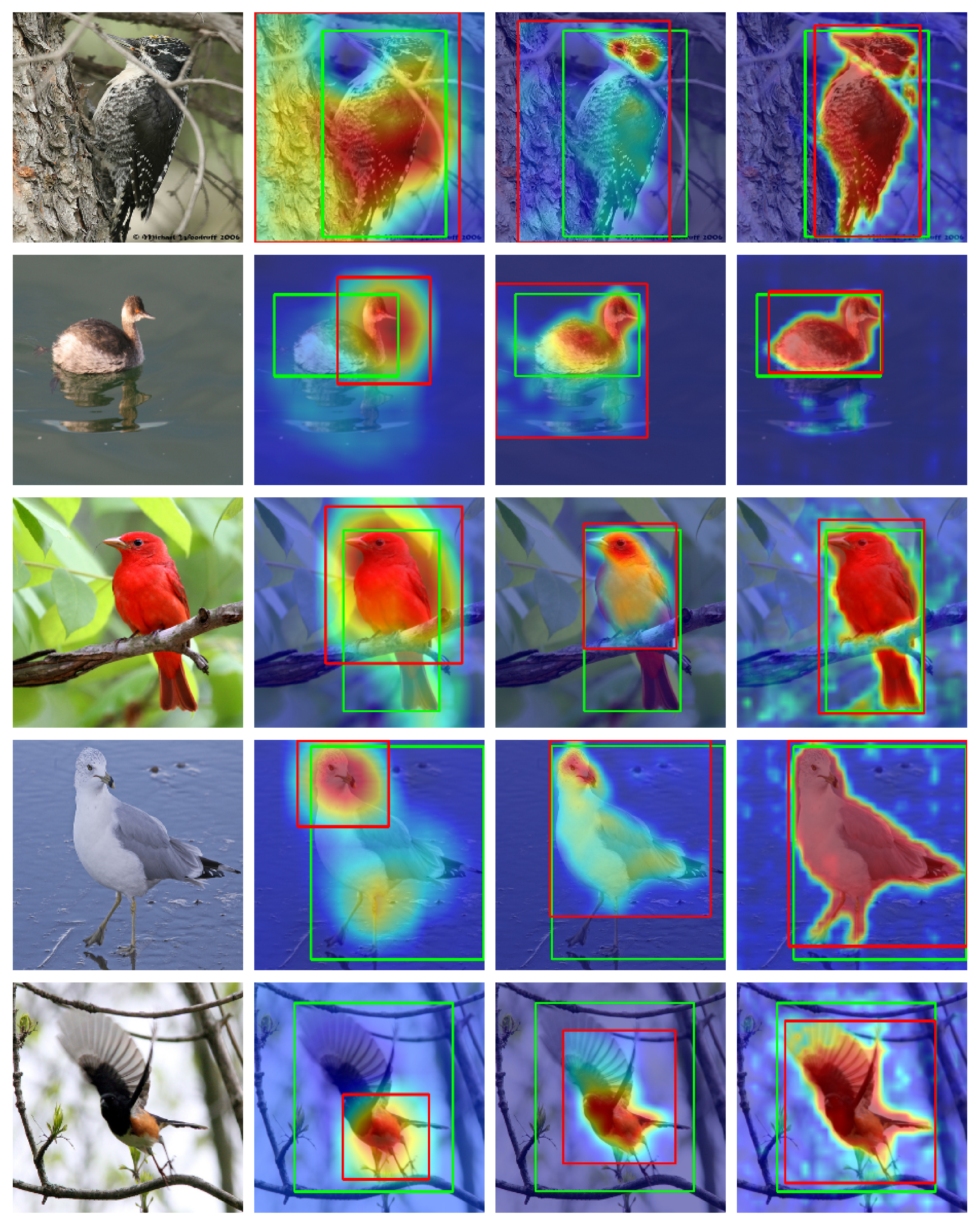}
    \end{minipage}
    \hspace{-0.01\linewidth}
    \begin{minipage}{0.485\linewidth}
        \centering
        \includegraphics[width=\linewidth]{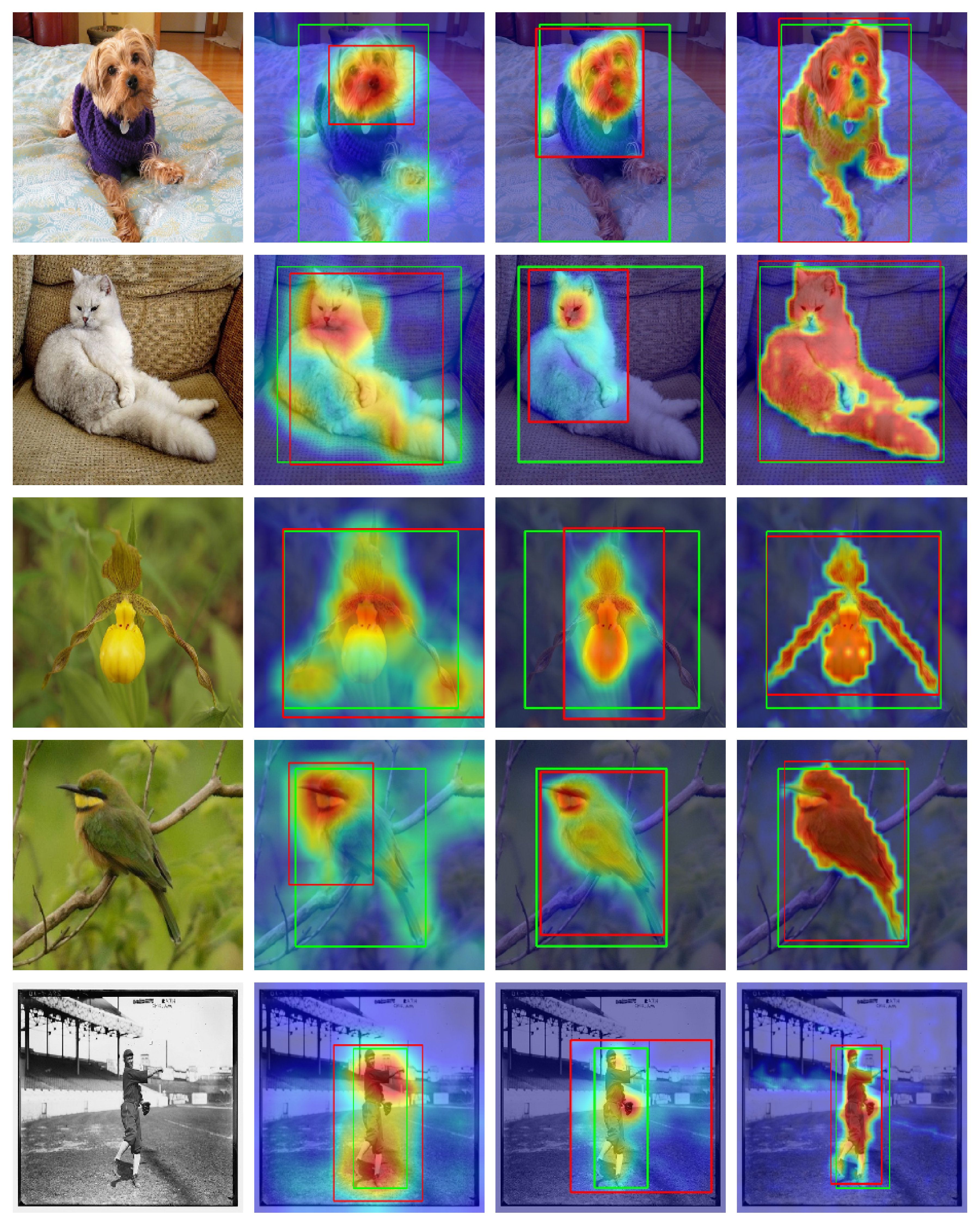}
    \end{minipage}

    \vspace{0.3em}

    \begin{minipage}{0.485\linewidth}
        \centering
        \small (a) CUB-200-2011
    \end{minipage}
    \hspace{-0.01\linewidth}
    \begin{minipage}{0.485\linewidth}
        \centering
        \small (b) ILSVRC
    \end{minipage}
    \caption{Comparison of localization results on CUB-200-2011 and ILSVRC datasets. Green and red colors are used for ground-truth and predicted bounding boxes.}
    \label{fig:dataset_comparison}
\end{figure*}

Weakly supervised object localization (WSOL) aims to localize object bounding boxes while using only image-level annotations for training. By removing the need for costly bounding box labels, WSOL can significantly reduce annotation costs. The pioneering work of Zhou \etal~\cite{zhou2016learning} introduces Class Activation Mapping (CAM), which localizes discriminative regions in a CNN trained purely for classification, has been widely adopted within several WSOL methods~\cite{kumar2017hide, zhang2018adversarial, choe2019attention, zhang2020rethinking, kim2022bridging}. However, this approach suffers from partial internal activation (Figure \ref{fig:dataset_comparison}), which leads to a bounding box prediction that fails to capture the entire extent of the object. Although many subsequent works have attempted to address this issue through spatial regularization~\cite{choe2019attention,wang122020minmaxcam, kim2022bridging}, input manipulation~\cite{kumar2017hide, yun2019cutmix}, disentangling localization from classification~\cite{zhang2020rethinking, xie2022contrastive}, or leveraging vision transformers~\cite{gao2021ts, chen2022lctr}, the challenge of incomplete activation still largely persists.

Recent methods such as GenPromp~\cite{xie2022contrastive} and C2AM~\cite{zhao2023generative} have shown promising results in WSOL, but often at the cost of multi-stage training and reliance on computationally heavy networks. For instance, GenPromp~\cite{zhao2023generative} achieves high performance by exploiting a vision-language model (CLIP) and Stable Diffusion while significantly increasing model size (1017M parameters). In this work, we propose \textbf{TriLite}, which employs a Vision Transformer (ViT) model pre-trained on the Large Vision Dataset (LVD-142M) as backbone and introduce \textit{TriHead}, a lightweight  module for localization. We freeze our ViT backbone and train only the TriHead module along with a single linear layer on the class token for classification. Freezing the backbone enables a single shared representation for both localization and classification, substantially reducing the number of trainable parameters and simplifying the overall model. In contrast, an unfrozen backbone introduces competing objectives for the two tasks, often necessitating separate networks to achieve comparable performance.

Unlike previous methods that split the input into two regions (foreground and background)~\cite{zhao2023generative, xu2022cream, xie2022contrastive, belharbi2022f}, the proposed TriHead module introduces an additional \textit{ambiguous} map. We hypothesize that certain image regions may belong to salient object(s) that are neither the primary target object nor the background. By assigning such regions to an ambiguous map, we reduce the noise that would otherwise occur if these regions were forced into binary categories.

Despite its simplicity, TriLite surpasses computationally demanding methods across established benchmarks, including CUB-200-2011~\cite{WahCUB_200_2011}, ImageNet-1K~\cite{ILSVRC15}, and OpenImages~\cite{benenson2019large, choe2020evaluating}. 
While CUB-200-2011 and ImageNet-1K are standard WSOL benchmarks, the inclusion of OpenImages further extends our evaluation to weakly supervised semantic segmentation (WSSS). 
Our main contributions are as follows:

\begin{itemize}
    \item We introduce the \textbf{TriHead} localization module, which can be integrated with existing pre-trained vision transformers, and complement it with a novel adversarial background loss that enhances object–background separation, a loss that, to the best of our knowledge, has not been previously explored in the WSOL literature.

    \item Our method is highly parameter-efficient, requiring fewer than 800K trainable parameters for ImageNet-1K, around 180K for CUB-200-2011, and around 90K for OpenImages—compared to existing methods that typically train with at least 22M parameters (Table~\ref{tab:wsol_comparison}).
    
    \item Unlike conventional WSOL methods that localize only sparse discriminative regions, our approach encourages complete object coverage (e.g., an entire dog rather than only its head; see Figure~\ref{fig:dataset_comparison}).
    
    \item TriLite achieves new state-of-the-art results in both WSOL and WSSS, with qualitative results demonstrating its ability to produce high-resolution, segmentation-like outputs.
\end{itemize}

\section{Related Work}
\label{sec:related_work}

A foundational method in WSOL is Class Activation Mapping (CAM)~\cite{zhou2016learning}, which leverages global average pooling (GAP) to produce localization maps from classification networks. Due to its  explainability nature, CAM focuses on the relevant features for the downstream problem, i.e. classification. Consequently, on the most discriminative regions, leading to incomplete or partial activation of the target object~\cite{choe2020evaluating}. 
To address this limitation, various strategies have been proposed, including spatial regularization~\cite{yun2019cutmix,choe2019attention,zhai2024background}, adversarial erasing~\cite{kumar2017hide,zhang2018adversarial}, and explicitly decoupling localization from classification~\cite{zhang2020rethinking,xie2022contrastive,kim2022bridging}.

Spatial regularization methods aim to distribute activations across broader object regions, thereby alleviating the partial localization problem. For instance, randomly masking~\cite{kumar2017hide} or mixing~\cite{yun2019cutmix} image patches encourages the model to attend to different object parts. Alternatively, instead of random masking, attention-based dropout in ADL~\cite{choe2019attention} and adversarial erasing in ACoL~\cite{zhang2018adversarial} mask highly activated regions, compelling the network to discover complementary object areas. In contrast, SPG~\cite{zhang2018self} exploits highly activated region by considering high- and low-confidence values in attention maps as pseudo masks to supervise lower-level features, thus enabling the discovery of medium-confidence regions. Similarly, F-CAM~\cite{belharbi2022f} derives pseudo labels directly from CAM activations and employs a decoder trained on top of the CAM encoder to recover high-resolution activation maps. Beyond spreading activations, Background Activation Suppression (BAS)~\cite{zhai2024background} differentiates foreground from background by measuring their contribution to the correct class and constraining the size of the foreground map. While our method also introduces a third undefined region similar to SPG~\cite{zhang2018self}, we differ in that this region is not defined by medium-confidence values and is not collapsed into either foreground or background. Compared to BAS~\cite{zhai2024background}, we explicitly disentangle classification from localization objectives, have a frozen backbone and make no assumptions about the extent of the foreground region.

An alternative research direction explicitly targets the misalignment between the localization and classification tasks inherent in WSOL. PSOL~\cite{zhang2020rethinking} introduced a two-stage approach separating classification and localization networks to reduce their conflicting objectives. Following this, C2AM~\cite{xie2022contrastive} proposed a dedicated disentangler module explicitly dividing features into binary foreground and background representations, effectively clarifying semantic regions. 
\cite{kim2022bridging} proposed objective alignment methods to further unify localization and classification branches, showing notable improvements in localization accuracy. Our method adopts a similar strategy to~\cite{zhang2020rethinking, xie2022contrastive} of decoupling localization and classification; however, instead of using separate networks, we employ two independent branches built on a shared backbone.

More recently, Vision Transformer (ViT) architectures have emerged as powerful tools in WSOL, given their capability to capture global context and long-range dependencies. TS-CAM~\cite{gao2021ts} integrated semantic-agnostic attention with class-specific maps to produce more comprehensive activation patterns. LCTR~\cite{chen2022lctr} utilized class-specific transformer pooling to leverage cross-patch relationships, enhancing the object localization performance. SCM~\cite{bai2022weakly} introduced self-complementary masks derived from transformer features, further discovering complementary object parts and significantly reducing localization ambiguity. In our work, we also leverage a Vision Transformer as the backbone for both localization and classification. However, unlike previous methods, we use a frozen ViT pre-trained in a self-supervised manner to have more universal features.

Complementary to these transformer-based WSOL methods, another line of work employs a frozen self-supervised ViT and applies post-hoc strategies for object localization without training on the target dataset. LOST~\cite{simeoni2021localizing} expands seed patches based on intra-image correlations, while TokenCut~\cite{wang2022self} formulates localization as a graph partitioning problem using normalized cuts. Similar to these methods, we also rely on a frozen self-supervised ViT backbone; however our approach introduces a lightweight learnable module for localization.

Beyond transformer-based approaches, recent work has explored generative and prompt-based modeling paradigms. GenPromp~\cite{zhao2023generative} formulated WSOL as a conditional image denoising problem, effectively combining generative and discriminative embeddings through pretrained CLIP~\cite{radford2021learning} models. Despite its state-of-the-art results, GenPromp requires large-scale generative models and complex multi-step training processes, imposing substantial computational costs.

Though significant progress has been made, challenges remain in efficiently addressing object regions and semantic misalignments with a minimal computational budget. Motivated by these gaps, our method, TriLite, introduces a straightforward yet powerful localization pipeline using a minimal parameter footprint.
\section{TriLite}
\label{sec:method}
\begin{figure*}
    \centering
    \includegraphics[width=0.94\linewidth]{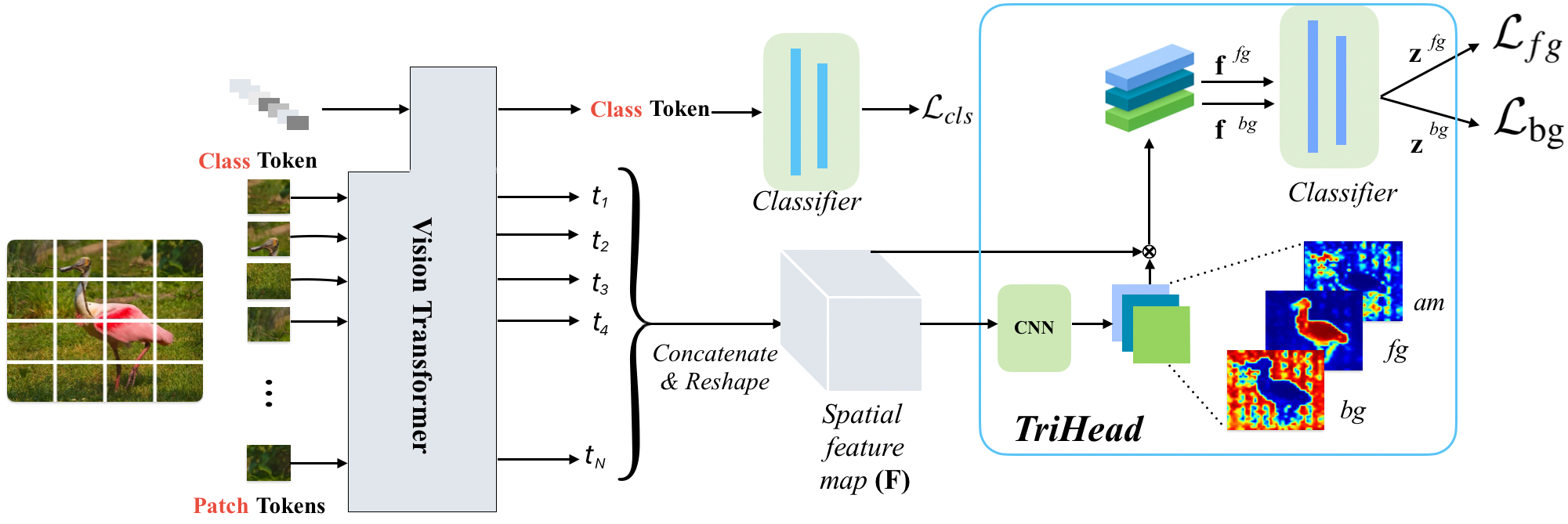}
    \caption{Overview of TriLite for WSOL. A frozen ViT backbone extracts patch features, while only a lightweight classification layer is trained using class token. The TriHead module applies a single convolutional layer to produce foreground, background, and ambiguous heatmaps. Supervision is applied to foreground and background embeddings.}
    \label{fig:architecture}
\end{figure*}

We build our framework on top of a frozen ViT-S/14 backbone, with Dinov2 pre-training~\cite{oquab2023dinov2}, which is used solely as a feature extractor. The architecture consists of two lightweight branches: one for classification and one for localization. The classification branch applies a single linear layer to the final class token produced by the Vision Transformer. For localization, we introduce a dedicated module called \emph{TriHead}, which operates independently of the classification pathway. An overview of the TriLite architecture is presented in Figure~\ref{fig:architecture}.

\textbf{Backbone.} To overcome the limited receptive fields of CNNs, we adopt Vision Transformers (ViTs) for their ability to capture global context. In particular, we use ViT-S/14 with DINOv2 pre-training~\cite{oquab2023dinov2} on the large-scale LVD-142M dataset (142M images) in a self-supervised setting. This training strategy produces general-purpose representations effective across both image-level (classification) and pixel-level (segmentation) tasks. Unlike prior WSOL approaches leveraging ViT backbones~\cite{gao2021ts, bai2022weakly}, we keep the backbone frozen throughout training, thereby preserving the semantically rich features learned via self-supervision. By contrast, supervised pre-training tends to bias features toward task-specific labels, limiting their semantic generality.


\noindent\textbf{TriHead Module.} 
Given the final ViT patch tokens $\mathbf{T} \in \mathbb{R}^{D \times N}$, where $D$ is the embedding dimension and $N = w \times h$ is the number of spatial patches, we reshape them into a feature map $\mathbf{F} \in \mathbb{R}^{D \times w \times h}$. The TriHead module processes $\mathbf{F}$ with a single convolutional layer, followed by batch normalization and a softmax activation:
\begin{equation}
\mathbf{M} = \text{Softmax}(\text{BN}(\text{Conv}(\mathbf{F}))), \quad \mathbf{M} \in \mathbb{R}^{3 \times w \times h},
\end{equation}


where $\mathbf{M} = [\mathbf{M}^{am}, \mathbf{M}^{fg},\mathbf{M}^{bg}]$ represents stack of ambiguous, foreground, and background maps.

Unlike conventional binary foreground–background separation, which forces every patch into one of two classes, our three-channel formulation allows the model to handle non-target salient regions more effectively. The effectiveness of this design choice is validated both quantitatively and qualitatively in Section~\ref{sec:ablation}.

Since softmax is applied across the three maps, supervising two maps is sufficient, as the third is implicitly constrained. Hence, during training we supervise only the foreground and background maps. To obtain compact feature representations, we compute a weighted average of the patch features $\mathbf{F}$ using the corresponding masks $\mathbf{M}^c$:
\begin{equation}
\mathbf{f}^{c} = \frac{\sum_{i=1}^{wh} \mathbf{M}_i^{c}\mathbf{F}_i}{\sum_{i=1}^{wh}\mathbf{M}_i^{c} + \epsilon}, \quad c \in \{fg, bg\},
\end{equation}
where $\epsilon$ is a small constant to avoid division by zero. The resulting vectors $\mathbf{f}^{fg}$ and $\mathbf{f}^{bg}$ represent the aggregated features of their respective regions, which are subsequently passed through a shared fully-connected layer to produce classification logits $\mathbf{z}^{fg}$ and $\mathbf{z}^{bg}$.

\noindent\textbf{Localization Loss.} 
To explicitly guide object localization, we supervise the foreground representation $\mathbf{z}^{fg}$ with the ground-truth class label $y$ using a standard cross-entropy loss:
\begin{equation}
\mathcal{L}_{fg} = \text{CrossEntropy}(\mathbf{z}^{fg}, y).
\end{equation}

While $\mathcal{L}_{fg}$ encourages correct classification from foreground regions, it does not explicitly discourage target-related features from leaking into the background map. 
To address this, we introduce an adversarial background loss that penalizes any activation of the target class in $\mathbf{z}^{bg}$:
\begin{equation}
\mathcal{L}_{bg} = -\log\left(1 - \frac{\exp(z^{bg}_{y})}{\sum_{j} \exp(z^{bg}_{j})} + \epsilon\right),
\end{equation}
where $z^{bg}_j$ denotes the background logit for class $j$, and $z^{bg}_y$ is the logit corresponding to the ground-truth class. 
This formulation encourages $\mathbf{M}^{bg}$ to exclusively activate in regions unrelated to the target object, thereby enhancing the separation between true background and discriminative object patches.


\noindent\textbf{Classification Branch.} 
For image-level classification, we attach a single fully-connected layer to the ViT class token $\mathbf{z}^{I}$ and supervise it with the ground-truth label $y$ using a standard cross-entropy loss:
\begin{equation}
\mathcal{L}_{cls} = \text{CrossEntropy}(\mathbf{z}^{I}, y).
\end{equation}

\noindent\textbf{Training Objective.} 
The overall objective combines localization and classification supervision:
\begin{equation}
\mathcal{L}_{total} = \mathcal{L}_{fg} + \alpha \mathcal{L}_{bg} + \mathcal{L}_{cls},
\end{equation}
where $\alpha$ is a weighting factor that balances the adversarial background loss relative to the other terms. 
Both the localization and classification branches are optimized jointly in a single-stage training scheme, avoiding multi-stage pipelines commonly used in prior WSOL approaches.

\section{Experiments}
\subsection{Experimental Settings}
\textbf{Datasets.} To ensure comparability with prior work in WSOL, we follow the established evaluation protocol and use the benchmark datasets defined in~\cite{choe2020evaluating}, which have become the standard criterion adopted by virtually all subsequent studies in this field.
We evaluate the proposed method on three widely adopted WSOL benchmarks: CUB-200-2011~\cite{WahCUB_200_2011}, ImageNet-1K~\cite{ILSVRC15} and OpenImages~\cite{benenson2019large, choe2020evaluating}. The CUB-200-2011 dataset comprises 11,788 images across 200 fine-grained bird species, split into 5,994 training and 5,794 test images. The ImageNet-1K dataset includes around 1.2 million images spanning 1,000 object categories, with 50,000 validation images used for testing. The OpenImages dataset used in our experiments contains 37,319 images across 100 classes, with 29,819 images for training, 2,500 for validation, and the remaining 5,000 for testing.

\noindent\textbf{Evaluation Metrics.} Following the standard WSOL evaluation protocol~\cite{choe2020evaluating}, we report Top-1 localization accuracy (Top-1 Loc), Top-5 localization accuracy (Top-5 Loc), and GT-known localization accuracy (GT Loc) on CUB-200-2011 and ImageNet-1K. Top-1 and Top-5 Loc measure localization success when (i) the classification result is correct within the top-1/top-5 predictions and (ii) the Intersection-over-Union (IoU) between the predicted bounding box and the ground truth exceeds 50\%. GT Loc evaluates localization accuracy independently of classification, by computing IoU using the provided ground-truth label. For segmentation performance on the OpenImages dataset, we adopt the Pixel-wise Average Precision (PxAP) metric proposed by~\cite{choe2020evaluating}, which measures the area under the precision–recall curve across all pixels.

\noindent\textbf{Implementation Details.}
To enable a more comprehensive evaluation, we implement our method with three variants of Vision Transformers. Our primary model uses a ViT-S/14 backbone with DINOv2 pre-training~\cite{oquab2023dinov2}. In addition, we evaluate ViT-S/16 with DINO pre-training~\cite{caron2021emerging} (self-supervised on ImageNet) and DeiT-S~\cite{pmlr-v139-touvron21a}, a supervised ViT-S/16 variant trained on ImageNet. It should be noted that all backbones are frozen during training.

The proposed TriHead module and the classification head are optimized using Adam. The classification branch is updated with a learning rate of $\text{base\_lr} \times \text{lr\_multiplier}$
, while the TriHead module is trained with the base learning rate. For CUB-200-2011, we train for 30 epochs with a batch size of 128, a base learning rate of $5 \times 10^{-5}$, and a learning rate multiplier of 10. For ImageNet-1K, training is performed for 20 epochs with a batch size of 256, a base learning rate of $5 \times 10^{-6}$, and a learning rate multiplier of 50. For OpenImages, we train for 30 epochs with a batch size of 32, using a learning rate of $1 \times 10^{-5}$ for both the classification and localization branches. Across all datasets, we apply a weight decay of 0.01 and employ early stopping with a patience of 5 epochs based on validation performance. The weighting factor $\alpha$ for the adversarial loss is determined via grid search on the validation set. For reproducibility, we provide full implementation and configuration details in our released code.

\begin{table*}[t]
    \centering
    \small
    \setlength{\tabcolsep}{6pt}
    \renewcommand{\arraystretch}{1.1}
    \begin{tabular*}{\textwidth}{@{\extracolsep{\fill}} l c c ccc ccc @{}}
        \toprule
        Method & Backbone & Params (M) & \multicolumn{3}{c}{CUB-200-2011} & \multicolumn{3}{c}{ImageNet-1K} \\
               &          &            & Top-1 & Top-5 & GT               & Top-1 & Top-5 & GT \\
        \midrule
        CAM (CVPR'16)~\cite{zhou2016learning}      & VGG-16                                  & 138        & 41.1 & 50.7 & 55.1 & 42.8 & 54.9 & 59.0 \\
   
        TS-CAM (ICCV'21)~\cite{gao2021ts}          & DeiT-S                                  & 22.1         & 71.3 & 83.8 & 87.7 & 53.4 & 64.3 & 67.6 \\
        DA-WSOL (CVPR'22)~\cite{zhu2022weakly}          & ResNet-50                              & 25.6& 66.7 & 81.83 & 69.9 & 55.8 & 70.3 & 68.23 \\
        LCTR (AAAI'22)~\cite{chen2022lctr}         & DeiT-S                                  & 22.1         & 79.2 & 89.9 & 92.4 & 56.1 & 65.8 & 68.7 \\
        SCM (ECCV'22)~\cite{bai2022weakly}         & DeiT-S                                  & 22.1         & 76.4 & 91.6 & 96.6 & 56.1 & 66.4 & 68.8 \\
        CREAM (CVPR'22)~\cite{xu2022cream}         & InceptionV3                             & 23.8         & 71.8 & 86.4 & 90.4 & 56.1 & 66.2 & 69.0 \\
        BAS (IJCV'24)~\cite{zhai2024background}    & ResNet-50                               & 25.6       & 76.8 & 90.0 & 95.4 & 57.5 & 68.6 & 72.0\\ 
        \midrule
        PSOL (CVPR'20)~\cite{zhang2020rethinking}  & \mbox{DenseNet-161 + EfficientNet-B7}   & \mbox{28.7 + 66} & 80.9 & 90.0 & 91.8 & 58.0 & 65.0 & 66.3 \\
        C2AM (CVPR'22)~\cite{xie2022contrastive}   & \mbox{DenseNet-161 + EfficientNet-B7}   & \mbox{28.7 + 66} & 81.8 & 91.1 & 92.9 & 59.6 & 67.1 & 68.5 \\
        GenPromp (ICCV'23)~\cite{zhao2023generative} & \mbox{Stable Diffusion + EfficientNet-B7} & \mbox{1017$^{\dagger}$ + 66} & 87.0 & 96.1 & 98.0 & 65.2 & 73.4 & 75.0 \\
        \midrule
        TriLite (ours) & DeiT-S &  22.1 + 0.8$^{\ddagger}$ & 69.2 & 88.2 & 95.1 & 59.2 & 69.1 & 71.7 \\
        TriLite (ours) & DINO &  22.1 + 0.8$^{\ddagger}$ & 76.3 & 91.1 & 95.9 & 58.4 & 69.7 & 73.2\\
        TriLite (ours)                                      & ViT-S/14 (DINOv2)                           & 22.1 + 0.8$^{\ddagger}$   & \textbf{87.3}
        & \textbf{96.7} & \textbf{98.5} & \textbf{65.5} & \textbf{75.6} & \textbf{77.9} \\
        \bottomrule
    \end{tabular*}
    \caption{Comparison with state-of-the-art WSOL methods on CUB-200-2011 and ImageNet-1K.
    $^{\dagger}$~GenPromp fine-tunes 898M parameters out of a total of 1017M. 
$^{\ddagger}$~TriLite freezes the 22.1M ViT-S/14 backbone and trains only 0.8M additional parameters.}
    \label{tab:wsol_comparison}
\end{table*}

\subsection{Results of WSOL} 
We compare TriLite against recent state-of-the-art WSOL approaches in Table~\ref{tab:wsol_comparison} and Table~\ref{tab:openimages}. 
Following prior works, we also provide qualitative comparisons against CAM~\cite{zhou2016learning} and the state-of-the-art method in Figure~\ref{fig:dataset_comparison}, with additional examples and a demonstration video included in the supplementary material. 

For our best-performing variant with DINOv2 pre-training, TriLite achieves state-of-the-art performance on ImageNet-1K, surpassing the previous best approach, GenPromp~\cite{zhao2023generative}, by \textbf{+0.3\%}, \textbf{+2.2\%}, and \textbf{+2.9\%} in Top-1, Top-5, and GT-known localization accuracy. 
Against other strong competitors such as C2AM~\cite{xie2022contrastive} and BAS~\cite{zhai2024background}, TriLite delivers even larger gains, underscoring its robustness on this challenging benchmark. 
On CUB-200-2011, TriLite also outperforms GenPromp by \textbf{+0.3\%}, \textbf{+0.6\%}, and \textbf{+0.5\%} in Top-1, Top-5, and GT-known localization accuracy, while clearly surpassing C2AM and BAS. These consistent improvements across both benchmarks highlight the effectiveness of TriLite and set a new state of the art in WSOL.

\subsection{Supervised vs. Self-supervised Backbones}
Considering the TriLite variants in Table~\ref{tab:wsol_comparison}, 
DeiT-S backbone demonstrate superior localization performance (GT-known localization accuracy) compared to prior studies employing the same architecture, despite keeping the backbone frozen. However, the classification accuracy remains competitive only on ImageNet-1K, where the backbone was originally trained, and shows a noticeable drop on CUB-200-2011, highlighting the limitations of supervised pre-training when transferring across datasets. This contrast becomes more apparent when comparing DeiT-S with its DINO counterpart, both trained on ImageNet-1K using the same architecture. Unlike supervised pre-training, which may bias the backbone features toward the pretraining dataset, DINO yields task-agnostic representations that generalize better. Consequently, both classification and localization performance improve, surpassing all prior methods except GenPromp, which relies on a significantly larger model.

\subsection{Analysis of CUB-200-2011 Performance} 
\begin{figure}[t]
  \centering
  \begin{overpic}[width=0.9\linewidth]{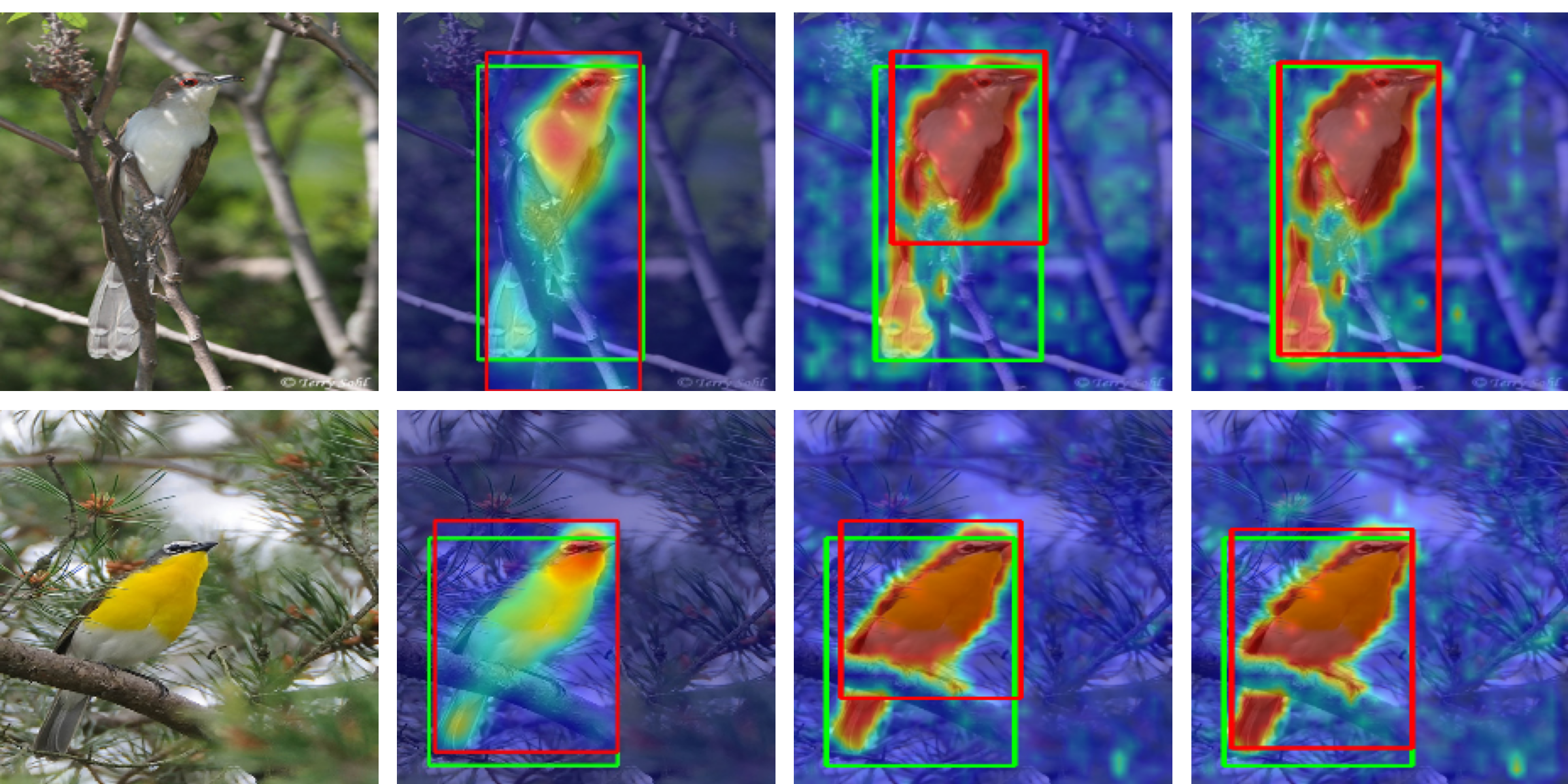}
    \put(12.5,52){\makebox(0,0){\small Image}}
    \put(37.5,52){\makebox(0,0){\small GenPromp}}
    \put(62.5,52){\makebox(0,0){\small Partial}}
    \put(87.5,52){\makebox(0,0){\small Merged}}
  \end{overpic}

\caption{Partial coverage on CUB-200-2011. TriLite activations may miss occluded regions, yielding fragmented bounding boxes (“Partial”). Merging all activated regions into a single box (“Merged”) recovers full coverage.}
  \label{fig:cub_occlusion}
\end{figure}
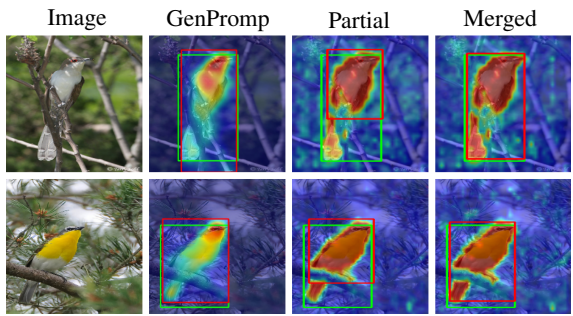

While TriLite achieves strong results on the CUB-200-2011 dataset, the performance margin over GenPromp~\cite{zhao2023generative} is smaller compared to ImageNet-1K. 
Upon closer examination, we attribute this behavior to combination of two factors: 
(i) the dataset primarily consists of close-up images of the target class, and 
(ii) unlike prior WSOL methods that often produce spatially diffuse activations, TriLite generates high-resolution and spatially precise activation maps. 
Consequently, when a target object is partially occluded (e.g., by branches), the occluded regions receive little or no activation, resulting in fragmented foreground regions and smaller bounding boxes that only partially cover the object. 
This observation is supported by qualitative examples (Figure~\ref{fig:cub_occlusion}) and further validated through a post-processing step that enforces a single bounding box encompassing all activated regions. 
With this adjustment, TriLite achieves improvements of \textbf{+0.6\%}, \textbf{+0.7\%}, and \textbf{+0.7\%} in Top-1, Top-5, and GT-known localization accuracy, respectively, compared to the reported results for CUB-200-2011 in Table~\ref{tab:wsol_comparison}. 

We do not report this post-processed variant in our main results to avoid imposing additional constraints and to ensure a fair comparison with prior studies. 
This analysis highlights that highly precise activation of the object of interest does not necessarily correspond to higher localization accuracy in the aforementioned scenarios.

\begin{table}[t]
\centering
\small
\begin{tabular}{@{}p{0.7\linewidth}c@{}}
\toprule
Method & PxAP (\%) \\
\midrule
CAM \cite{zhou2016learning} (CVPR'16) & 58.0 \\
HaS \cite{kumar2017hide} (ICCV'17) & 58.2 \\
ACoL \cite{zhang2018adversarial} (CVPR'18) & 57.8 \\
SPG \cite{zhang2018self} (ECCV'18) & 57.7 \\
ADL \cite{choe2019attention} (CVPR'19) & 54.3 \\
CutMix \cite{yun2019cutmix} (ECCV'19) & 58.7 \\
CAM* + F-CAM (WACV'22) \cite{belharbi2022f} & 72.1 \\
DA-WSOL \cite{zhu2022weakly} (CVPR'22) & 65.42 \\
BAS \cite{zhai2024background}(IJCV'24) & 66.86 \\
TriLite (ours) & \textbf{73.3} \\
\bottomrule
\end{tabular}
\caption{PxAP (\%) on OpenImages dataset. CAM* is reproduced by F-CAM.}
\label{tab:openimages}
\end{table}

\subsection{Results of WSSS} 
\begin{figure}[t]
  \centering
  \begin{overpic}[width=0.9\linewidth]{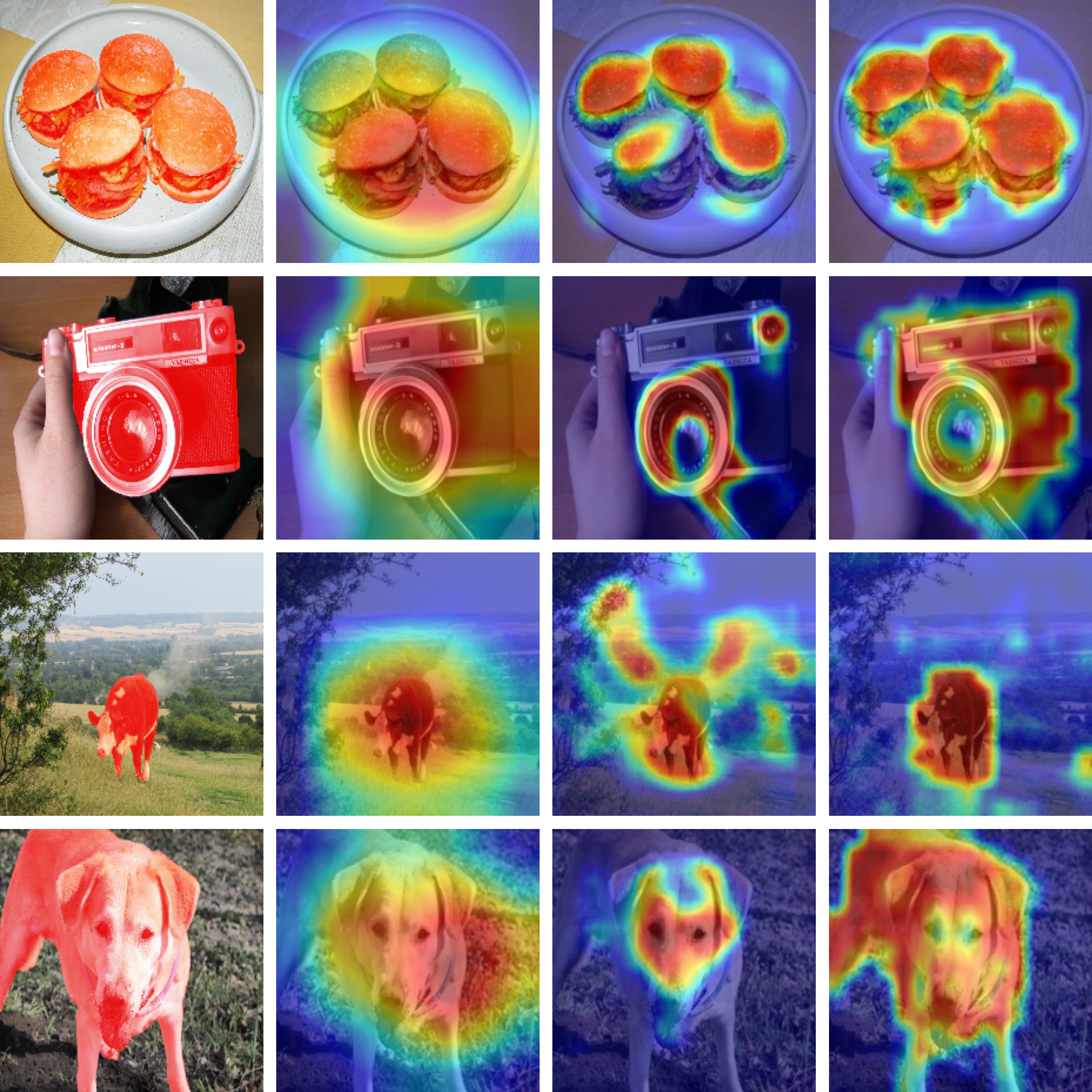}
   \put(12,102){\makebox(0,0){\small Image + Mask}}
    \put(37.5,102){\makebox(0,0){\small CAM}}
    \put(62.5,102){\makebox(0,0){\small BAS}}
    \put(87.5,102){\makebox(0,0){\small Ours}}

  \end{overpic}

\caption{Qualitative results on the OpenImages dataset. The first column shows the original image with the ground-truth mask overlaid.}

  \label{fig:openimages}
\end{figure}
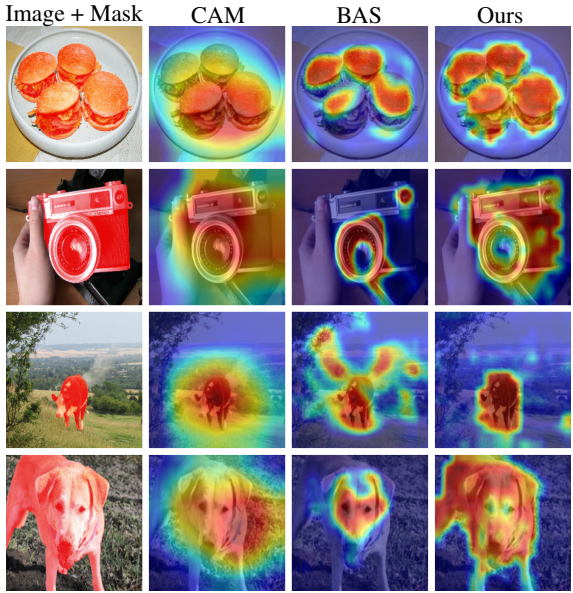

To extend beyond bounding-box localization and follow the evaluation protocol of~\cite{choe2020evaluating}, we report pixel-level accuracy on the OpenImages dataset. 
OpenImages is a particularly challenging benchmark due to its high intra-class variation~\cite{belharbi2022f} and complex backgrounds~\cite{zhai2024background}. 
Consequently, many recent methods have not reported results on this dataset, and we therefore include comparisons with both traditional and more recent approaches. 
On OpenImages, TriLite establishes a new state of the art, surpassing F-CAM~\cite{belharbi2022f}, which has been the strongest reported method to date, as shown in Table~\ref{tab:openimages}.
A qualitative comparison is provided in Figure~\ref{fig:openimages}, illustrating representative segmentation results from TriLite and competing methods. Since F-CAM does not publicly release trained weights for qualitative evaluation, we provide visual comparisons against BAS~\cite{zhai2024background}, the strongest available method after F-CAM.

\subsection{Parameter Efficiency and Simplicity} 
A key advantage of our approach is its parameter efficiency and streamlined training pipeline. 
Recent state-of-the-art WSOL methods often rely on multi-stage training~\cite{xie2022contrastive, zhang2020rethinking, zhao2023generative} or employ computationally demanding architectures~\cite{zhao2023generative}. 
For instance, GenPromp~\cite{zhao2023generative} achieves comparable localization performance to our method (Table~\ref{tab:wsol_comparison}), but requires approximately 1 billion trainable parameters and training on 8 RTX3090 GPUs. 
In contrast, TriLite leverages a frozen ViT-S/14 backbone and introduces only lightweight trainable components, resulting in fewer than \textbf{800K trainable parameters on ImageNet-1K}. 
The parameter count further scales linearly with the number of classes, requiring fewer than 1K additional parameters per class.

TriLite therefore achieves state-of-the-art localization accuracy while drastically reducing both model size and training requirements compared to prior approaches. 
Its single-stage training procedure makes the framework scalable across datasets and applications, providing a practical and accessible solution for the WSOL community.


\subsection{Ablation Study}
\label{sec:ablation}
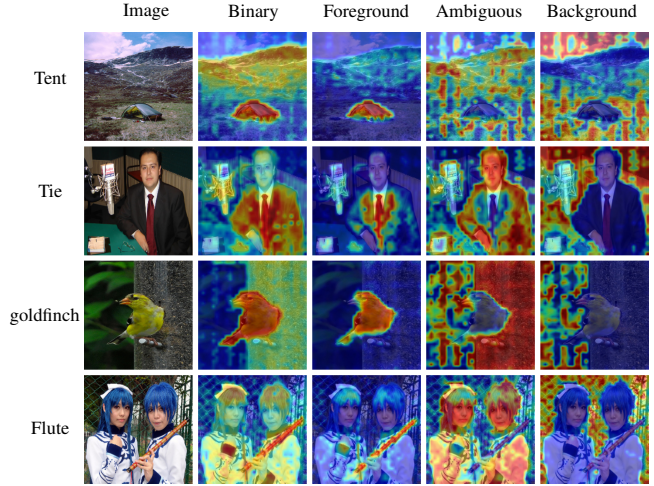
\begin{figure}[t]
  \centering
  \begin{overpic}[width=0.9\linewidth]{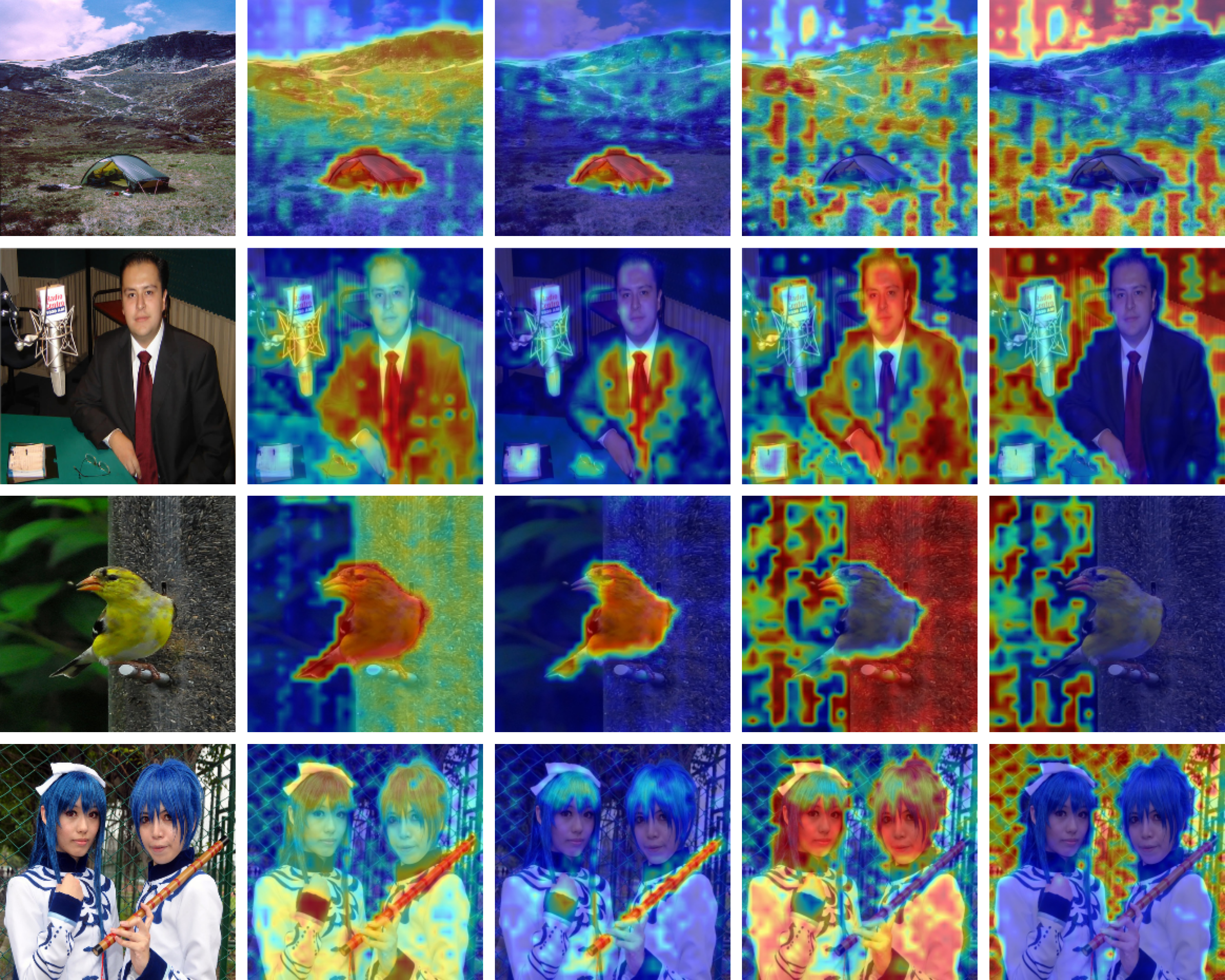}
    \put(-6,73){\makebox(0,0)[t]{\scriptsize Tent}}
    \put(-6,53){\makebox(0,0)[t]{\scriptsize Tie}}
    \put(-7,31){\makebox(0,0)[t]{\scriptsize goldfinch}}
    \put(-6,11){\makebox(0,0)[t]{\scriptsize Flute}}

    \put(11,85){\makebox(0,0)[t]{\scriptsize Image}}
    \put(30,85){\makebox(0,0)[t]{\scriptsize Binary}}
    \put(50,85){\makebox(0,0)[t]{\scriptsize Foreground}}
    \put(70,85){\makebox(0,0)[t]{\scriptsize Ambiguous}}
    \put(90,85){\makebox(0,0)[t]{\scriptsize Background}}
  \end{overpic}
\caption{Comparison between binary and three-channel outputs. 
Regions mistakenly activated in the binary setting are reassigned to the ambiguous channel in the three-channel formulation.}

  \label{fig:multi_vs_binary}
\end{figure}

Our ablation study is divided into two parts: (i) validation of the three-channel output and the impact of the adversarial loss, and (ii) assessment of the effectiveness of the TriHead module. 
Table~\ref{tab:ablation1} evaluates the effect of binary versus three-channel outputs, each with or without the adversarial loss. 
We find that neither the three-channel head alone nor the adversarial loss alone provides substantial improvement over the binary baseline. 
However, when combined, the three-channel output and adversarial loss yield a notable performance boost, achieving the best results on both CUB-200-2011 and ImageNet-1K. 
Qualitative comparisons in Figure~\ref{fig:multi_vs_binary} further illustrate this behavior: binary separation often produces extended foreground regions that include background elements (e.g., mountains, tree trunks) or non-target objects (e.g., suits, people), whereas in our approach the background map is encouraged to contain only non-salient patches and the ambiguous map captures the remaining non-target regions.

To evaluate the specific contribution of the TriHead module, we compare TriLite against WSOL methods that also rely on a frozen transformer backbone. 
Since no prior work has employed a vision transformer with DINOv2 pre-training, we consider methods based on the earlier DINO framework, which is likewise trained in a self-supervised manner. 
Accordingly, we train TriLite with a ViT-S/16 backbone using DINO pre-training and compare it to LOST~\cite{simeoni2021localizing} and TokenCut~\cite{wang2022self}, both of which adopt the same backbone. 
As all methods use a single linear layer over the class token for classification, differences in performance can be attributed entirely to the localization component. 
Under this controlled setting, TriLite achieves a clear improvement over both LOST and TokenCut, as shown in Table~\ref{tab:ablation2}. 
The results for LOST are taken directly from the original paper, while TokenCut is reproduced in our implementation, yielding slightly higher performance than originally reported.

\begin{table}[t]
    \centering
    \small
    \resizebox{\columnwidth}{!}{%
    \begin{tabular}{@{}l c ccc ccc@{}}
        \toprule
        Heatmap Type & Adv. Loss & \multicolumn{3}{c}{CUB-200-2011} & \multicolumn{3}{c}{ImageNet-1K} \\
        \cmidrule(lr){3-5} \cmidrule(l){6-8}
        & & Top-1 & Top-5 & GT & Top-1 & Top-5 & GT \\
        \midrule
        Binary         & \xmark & 86.7 & 95.8 & 97.6 & 64.6 & 74.3 & 76.5 \\
        Binary         & \cmark & 86.5 & 95.7 & 97.5 & 65.2 & 75.0 & 77.2 \\
        Three-channel  & \xmark & 85.0 & 95.0 & 97.0 & 65.0 & 75.0 & 77.4 \\
        Three-channel  & \cmark & \textbf{87.3} & \textbf{96.7} & \textbf{98.5} & \textbf{65.5} & \textbf{75.6} & \textbf{77.9} \\
        \bottomrule
    \end{tabular}%
    }
    \caption{Ablation study of the TriHead module. We compare heatmap types (binary vs. three-channel) and evaluate the effect of the adversarial loss. All experiments are conducted with ViT-S/14 (DINOv2 pre-training) combined with TriHead.}
    \label{tab:ablation1}
\end{table}


\begin{table}[t]
    \centering
    \small
    \resizebox{\columnwidth}{!}{%
    \begin{tabular}{@{}l c ccc ccc@{}}
        \toprule
        Method & Pre-training & \multicolumn{3}{c}{CUB-200-2011} & \multicolumn{3}{c}{ImageNet-1K} \\
        \cmidrule(lr){3-5} \cmidrule(l){6-8}
        & & Top-1 & Top-5 & GT & Top-1 & Top-5 & GT \\
        \midrule
        LOST (BMVC'21)~\cite{simeoni2021localizing} & DINO & 71.3 & — & 89.7 & 49.0 & - & 60.0 \\
        TokenCut (CVPR'22)~\cite{wang2022self} & DINO &  73.4 & 87.5 & 91.1 & 54.4 & 63.9 & 66.8 \\
        TriLite (ours) & DINO & \textbf{76.3} & \textbf{91.1} & \textbf{95.9} & \textbf{58.4} & \textbf{69.7} & \textbf{73.2}\\
        \bottomrule
    \end{tabular}%
    }
    \caption{Ablation on impact of TriHeade module. All methods use frozen ViT-S/16 with DINO pre-training as backbone.}
    \label{tab:ablation2}

\end{table}

\begin{figure}[t]
    \centering

    \begin{subfigure}{0.95\linewidth}
        \includegraphics[width=\linewidth]{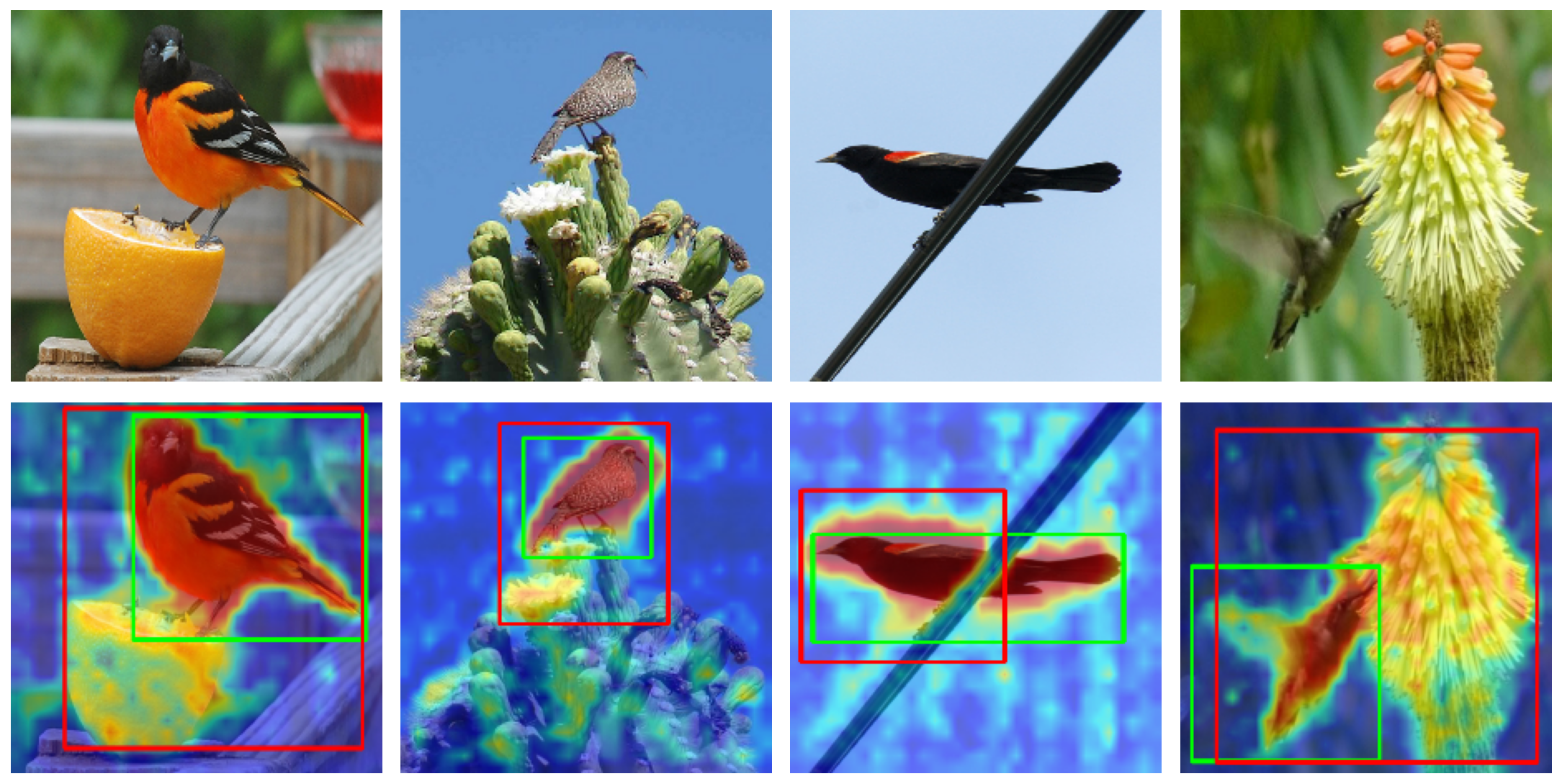}
        \caption{}
        \label{fig:failure_cub}
    \end{subfigure}

    \begin{subfigure}{0.95\linewidth}
        \includegraphics[width=\linewidth]{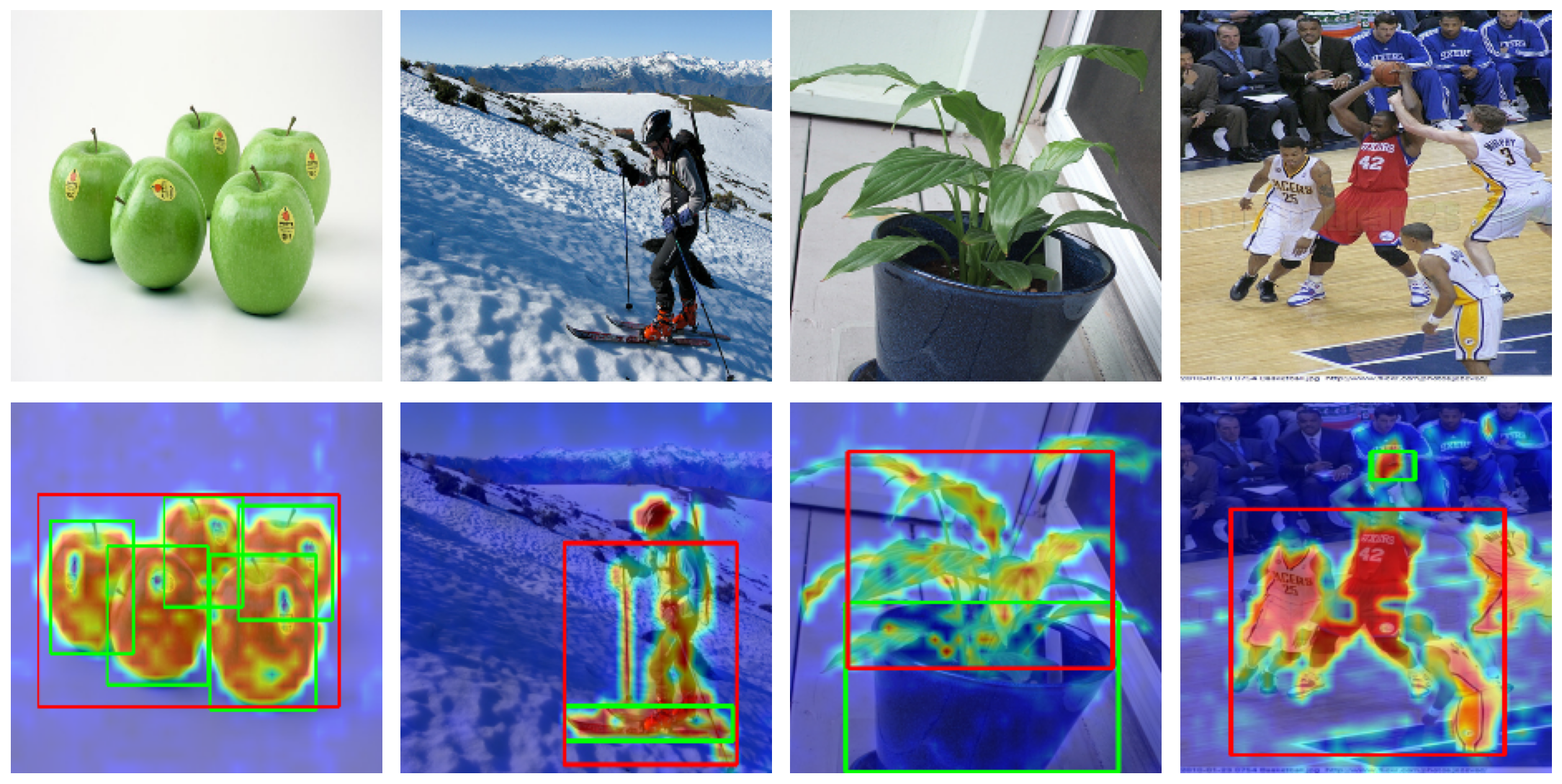}
        \caption{}
        \label{fig:failure_ILSVRC}
    \end{subfigure}

    \begin{subfigure}{0.95\linewidth}
        \includegraphics[width=\linewidth]{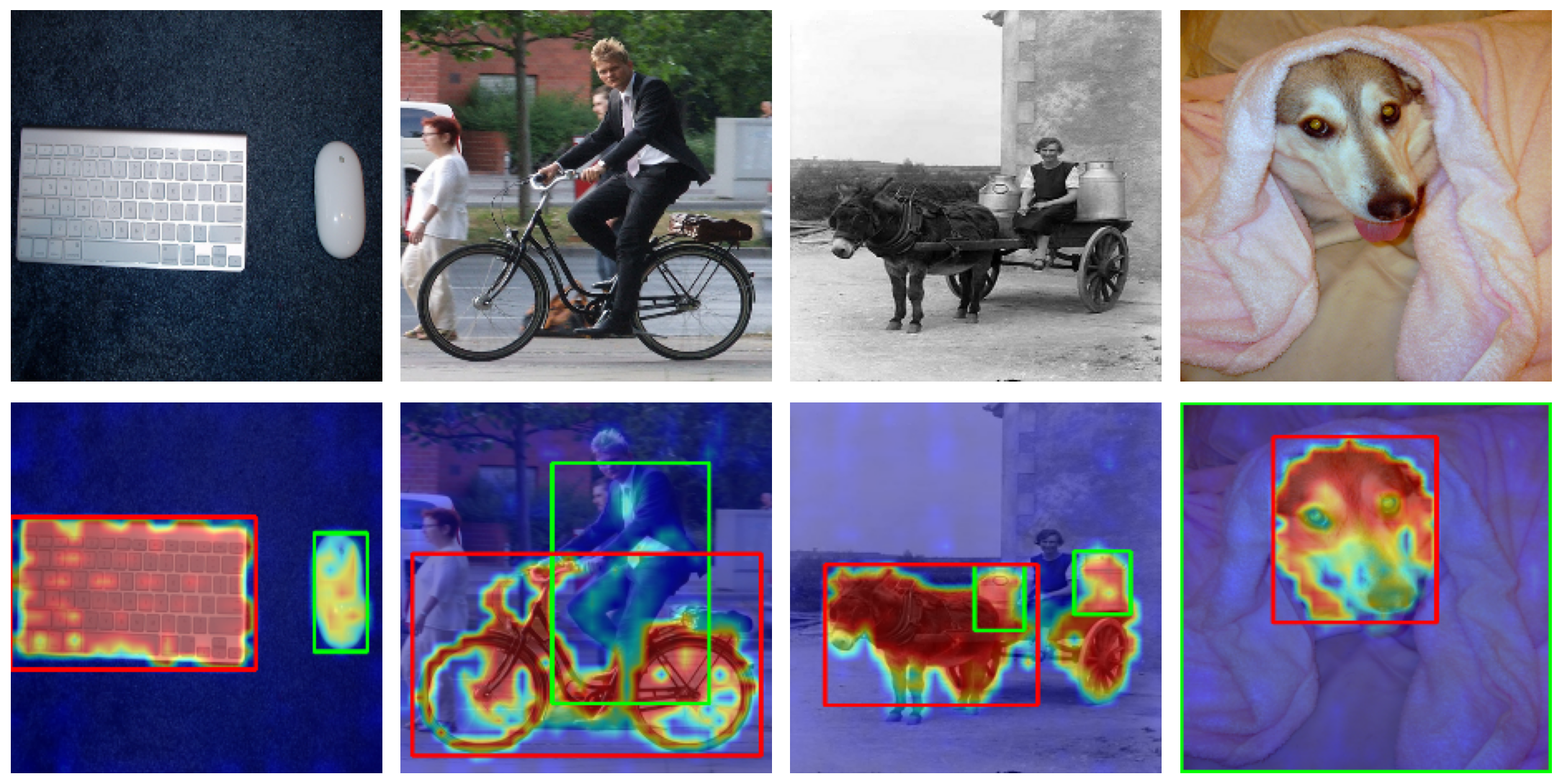}
        \caption{}
        \label{fig:failure_ILSVRC_misclassification}
    \end{subfigure}

    \makebox[\linewidth]{%
        \begin{tabular}{@{}*{4}{p{0.22\linewidth}@{\hspace{0.005\linewidth}}}@{}}
            \centering\textcolor{green}{\scriptsize Mouse} \\[-0.3em] \centering\textcolor{red}{\scriptsize Keyboard} &
            \centering\textcolor{green}{\scriptsize Suit} \\[-0.3em] \centering\textcolor{red}{\scriptsize Bicycle} &
            \centering\textcolor{green}{\scriptsize Milk Can} \\[-0.3em] \centering\textcolor{red}{\scriptsize Horse Cart} &
            \centering\textcolor{green}{\scriptsize Bath Towel} \\[-0.3em] \centering\textcolor{red}{\scriptsize Husky}
        \end{tabular}%
    }

    \caption{Examples of failure cases. (a) Mislocalization on CUB-200-2011. 
    (b) Mislocalization on ILSVRC. 
    (c) Mislocalization on ILSVRC due to misclassification.  GT class is shown in \textcolor{green}{green} and predicted class in \textcolor{red}{red}.}
    \label{fig:mis_localization}
\end{figure}

\subsection{Failure cases and Future Directions}
Although our method achieves state-of-the-art performance, analyzing failure cases is essential for guiding further improvements. On CUB-200-2011 (Figure~\ref{fig:failure_cub}), errors often stem from fragmented activations caused by occlusion or from highlighting salient but non-target regions that are nonetheless discriminative for the class. On ILSVRC (Figures~\ref{fig:failure_ILSVRC}, \ref{fig:failure_ILSVRC_misclassification}), three main failure modes are observed: (i) multi-instance scenarios (e.g., apples) where activations of several instances merge into a single bounding region, (ii) context-driven activations (e.g., basketball, skis, pot) where large contextual areas are emphasized instead of the primary object, and (iii) misclassification-induced errors, often arising when multiple object classes are present but only a single class label is provided. These cases complicate both classification and localization.

Addressing these limitations suggests two main directions. First, because TriLite decouples classification and localization, the predicted maps are class-agnostic, unlike CAM-based approaches that produce category-specific maps~\cite{zhang2020rethinking, xie2022contrastive}. While sufficient for single-object datasets, this design restricts performance in multi-class scenarios, where multiple categories may be activated without explicit control. Extending TriLite to generate class-specific localization maps, while avoiding over-reliance on discriminative regions, would improve applicability to multi-class benchmarks and real-world tasks. Second, improving continuity in multi-instance settings—where a single bounding box often covers multiple objects of the same class—would enhance the usability of localization outputs. Tackling these challenges would broaden the scope of WSOL beyond single-class assumptions and enable wider adoption in practical applications.

\section{Conclusion}
We introduced \textbf{TriLite}, a weakly supervised object localization framework that couples a frozen Vision Transformer (ViT) pretrained with self-supervision (DINOv2) with a lightweight \textit{TriHead} module. 
By training fewer than 800K parameters on ImageNet-1K, TriLite reconciles the localization–classification conflict through separate heads for each task. Extensive experiments on CUB-200-2011, ImageNet-1K, and OpenImages demonstrate that this minimalist design achieves new state-of-the-art performance in both WSOL and WSSS, while leveraging the semantic richness of pretrained features without costly end-to-end optimization. Future directions include extending the framework to better handle multi-instance scenarios, where multiple objects of the same class should be localized separately, and developing class-specific localization mechanisms for multi-class images. 

\section{Acknowledgements}
This work is partially funded by the DEFRA AHOI project, funded by the Belgian Royal Higher Institute for Defence, under contract number 23DEFRA002.

{
    \small
    \bibliographystyle{ieeenat_fullname}
    \bibliography{main}
}


\end{document}